\documentclass{article} 
\usepackage[final]{colm2025_conference}

\usepackage{microtype}
\definecolor{redlinkcolor}{rgb}{0.79607843, 0.25098039, 0.25882353}
\definecolor{bluecitecolor}{rgb}{0,0.36,0.69}
\usepackage[colorlinks=true,linkcolor=redlinkcolor,citecolor=bluecitecolor,urlcolor=bluecitecolor]{hyperref}
\usepackage{url}
\usepackage{booktabs}
\usepackage{wrapfig}
\usepackage{lineno}

\usepackage{tcolorbox}
\usepackage{graphicx}
\usepackage{xspace}
\usepackage[table]{xcolor}
\newcommand{\benchmark}{\textsc{ChemRAG-Bench}\xspace}
\newcommand{\toolkit}{\textsc{ChemRAG-Toolkit}\xspace}
\newcommand{\chemrag}{\textsc{ChemRAG}\xspace}

\usepackage{multicol}
\usepackage{multirow}
\usepackage{enumitem}

\title{Benchmarking Retrieval-Augmented Generation for Chemistry}

\author{Xianrui Zhong$^1$, Bowen Jin$^1$, Siru Ouyang$^1$, Yanzhen Shen$^1$, Qiao Jin$^2$, \\\textbf{Yin Fang$^2$, Zhiyong Lu$^2$ \& Jiawei Han$^1$} \\
$^1$Siebel School of Computing and Data Science, University of Illinois Urbana-Champaign \\
$^2$National Library of Medicine, National Institutes of Health \\
\texttt{\{xzhong23, bowenj4, siruo2, yanzhen4, hanj\}@illinois.edu} \\
\texttt{\{qiao.jin, yin.fang, zhiyong.lu\}@nih.gov}
}

\begin{document}

\ifcolmsubmission
\linenumbers
\fi

\maketitle

\begin{abstract}
Retrieval-augmented generation (RAG) has emerged as a powerful framework for enhancing large language models (LLMs) with external knowledge, particularly in scientific domains that demand specialized and dynamic information. 
Despite its promise, the application of RAG in the chemistry domain remains underexplored, primarily due to the lack of high-quality, domain-specific corpora and well-curated evaluation benchmarks. 
In this work, we introduce \benchmark, a comprehensive benchmark designed to systematically assess the effectiveness of RAG across a diverse set of chemistry-related tasks. 
The accompanying chemistry corpus integrates heterogeneous knowledge sources, including scientific literature, the PubChem database, PubMed abstracts, textbooks, and Wikipedia entries. 
In addition, we present \toolkit, a modular and extensible RAG toolkit that supports five retrieval algorithms and eight LLMs. 
Using \toolkit, we demonstrate that RAG yields a substantial performance gain—achieving an average relative improvement of 17.4\% over direct inference methods. 
We further conduct in-depth analyses on retriever architectures, corpus selection, and the number of retrieved passages, culminating in practical recommendations to guide future research and deployment of RAG systems in the chemistry domain. The code and data is available at \url{https://chemrag.github.io}.
\end{abstract}

\section{Introduction}

Retrieval-augmented generation (RAG) \citep{gao2023retrieval} has emerged as a powerful paradigm for enhancing large language models (LLMs) with external knowledge sources. 
By incorporating retrieval into the generation process, RAG can effectively mitigate hallucinations \citep{zhang2023siren} and inject up-to-date domain-specific information into LLMs \citep{siriwardhana2023improving}. 
These capabilities are particularly valuable in scientific domains, where factual accuracy and timely knowledge are critical. 
A typical scientific RAG system consists of two components: (1) a retriever that selects relevant documents or facts from a scientific knowledge base, and (2) a generator, often an LLM, that integrates the retrieved content to produce informed and coherent responses. 
Such frameworks have shown promising applications in domains like biomedicine \citep{xiong2024benchmarking}.

\textit{In chemistry}, LLMs have shown remarkable potential across various tasks, including molecular captioning~\citep{li2024empowering}, chemical reasoning~\citep{tang2025chemagent}, and reaction prediction~\citep{DBLP:conf/emnlp/Shi0ZL023}. However, chemistry is a highly specialized and dynamic discipline, characterized by complex terminologies, domain-specific conventions, and rapidly-evolving knowledge. As a result, LLMs trained on general corpora often fail to generate grounded and accurate responses, instead producing hallucinated or outdated content~\citep{DBLP:journals/corr/abs-2309-01219, wang2024knowledge}. RAG presents a natural solution to these limitations, allowing models to retrieve and incorporate trusted chemical knowledge during inference.

\begin{figure*}[t]
    \centering
    \includegraphics[width=\textwidth]{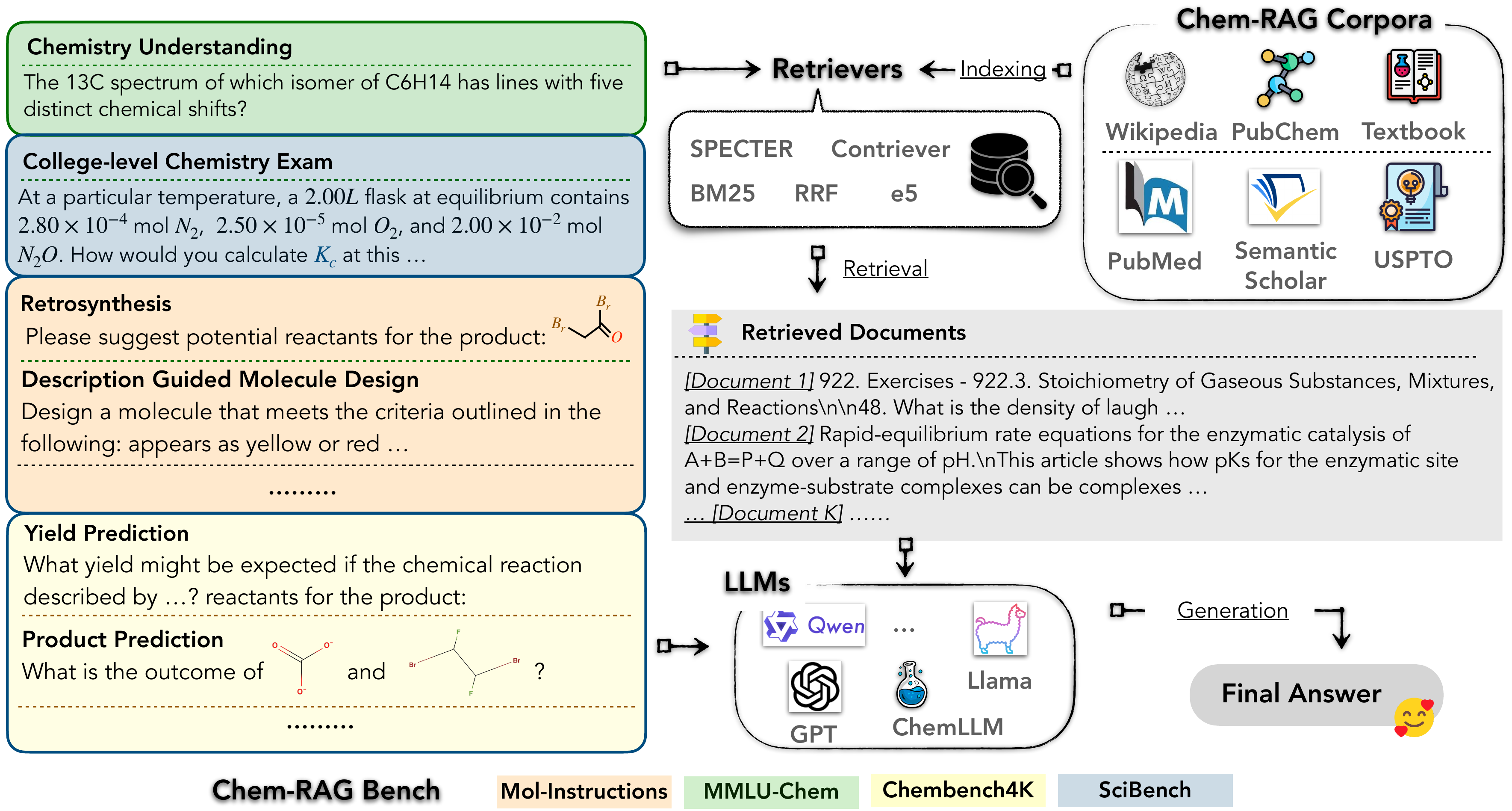}
    \caption{Overview of the \chemrag toolkit. Retrievers are first constructed from \chemrag corpora. For each question in the \benchmark benchmark, we retrieve related documents as additional contexts for LLMs to predict the final answer.}
    \label{fig:overview}
    \vspace{-5mm}
\end{figure*}

Despite the growing interest in applying RAG to the chemistry domain, there remains a lack of standardized benchmarks and curated domain-specific resources to support rigorous evaluation and design of RAG systems. To address this gap, we introduce \benchmark, a novel evaluation benchmark comprising 1,932 expert-curated question-answer pairs covering diverse chemistry tasks. These include description-guided molecular design, retrosynthesis, chemical calculations, molecule captioning, name conversion, and reaction prediction. This benchmark provides a foundation for systematically evaluating the effectiveness of RAG systems in chemistry and guiding future research in this direction.

To facilitate comprehensive and reproducible evaluation on \benchmark, we introduce \toolkit, a user-friendly and extensible toolkit that supports $6$ chemistry-related corpora, $5$ retrieval methods, and $8$ LLMs, encompassing both general-purpose and domain-specific LLMs.
Based on the proposed \benchmark benchmark, we conduct a systematic evaluation of various \chemrag solutions and analyze the impact of individual components on overall performance from multiple perspectives. 
Across a range of LLMs, we observe an average relative performance improvement of 17.4\% when using \chemrag compared to direct inference without retrieval. 

Along the retrieval corpus dimension, we find that different chemistry tasks exhibit distinct preferences for specific corpora. 
\textit{For instance, molecule design and reaction prediction tasks benefit more from literature-derived corpora, while nomenclature and conversion tasks favor structured chemical databases.} 
These observations suggest that task-aware corpus selection is crucial for maximizing RAG performance. Moreover, we show that combining all available corpora often yields the most robust results, serving as a comprehensive retrieval base.
In terms of the retriever component, Contriever \citep{izacard2021unsupervised} demonstrates consistently strong performance across tasks. We further find that performance can be enhanced by leveraging ensemble retrieval strategies that combine the strengths of multiple retrievers. 
Beyond standard evaluation metrics, we uncover a log-linear scaling trend between the number of retrieved passages and downstream model performance, indicating that retrieval depth plays a key role in generation quality. 
Additionally, we investigate the proportion of external knowledge utilized per task and provide in-depth analysis on retriever selection for chemistry discovery scenarios. 
Finally, we distill a set of practical recommendations from our findings, offering actionable insights for deploying and advancing RAG systems in the chemistry domain.
In summary, our key contributions are threefold:
\vspace{-2mm}
\begin{itemize}[leftmargin=*]
    \item We introduce \benchmark, a comprehensive benchmark comprising 1,932 expert-curated question-answer pairs across six chemistry-related knowledge sources, enabling systematic evaluation of RAG methods in the chemistry domain.
    \item We develop \toolkit, an easy-to-use and extensible framework that integrates five retrieval algorithms and eight large language models, and demonstrate an average relative improvement of 17.4\% when applying \chemrag over direct inference.
    \item We conduct comprehensive empirical analyses to examine the impact of retrieval corpus selection, retriever architecture, the number of retrieved documents, etc. Based on our findings, we provide practical guidelines to inform future research and the real-world deployment of chemistry-focused RAG systems.

\end{itemize}





\section{Related Work}

\subsection{Retrieval-Augmented Generation}
Retrieval-augmented generation (RAG) enhances large language models by incorporating external knowledge sources \citep{lewis2020retrieval}. 
It has been shown to reduce hallucinations \citep{Ayala_2024} and provide access to up-to-date information \citep{ragsurvey}. 
Recent work has sought to improve RAG performance through various enhancements, including more effective retrieval mechanisms \citep{glass2022re2g}, iterative retrieval-and-reasoning pipelines \citep{trivedi2022interleaving,jin2025search}, and the integration of long-context language models to better handle extended inputs \citep{jin2024long}.
While substantial progress has been made in general-domain RAG benchmarks \citep{asai2023selfraglearningretrievegenerate,yu2024autoragautonomousretrievalaugmentedgeneration,kwiatkowski2019natural,yang2018hotpotqa}, relatively little attention has been given to the scientific domain. 
Although recent efforts, such as \citet{xiong2024benchmarking}, begin to explore this direction in the medicine domain, the application of RAG to the chemistry domain remains underdeveloped. 
Notably, TextReact \citep{qian2023predictivechemistryaugmentedtext} applies text retrieval to tasks like reaction condition recommendation and one-step retrosynthesis. 
ChemLit-QA \citep{wellawatte2024chemlitqa} introduces a dataset for chemistry-oriented RAG, but its questions are generated from isolated paper excerpts and may lack real-world utility. 
Importantly, there remains a gap in the availability of high-quality, domain-specific corpora and comprehensive RAG benchmarks tailored to chemistry.

\subsection{Large Language Models for Chemistry}
The rapid advancement of large language models (LLMs) has opened up new opportunities across various scientific domains~\citep{ouyang2023the}, spurring the development of numerous benchmarks~\citep{lu2022learn, wang2023scibench, chembench4k}. 
Among these domains, chemistry stands out as a particularly challenging yet promising area for LLM applications~\cite{fang2023mol}. 
Recent efforts, such as ChemCrow~\citep{bran2023chemcrow}, have demonstrated the potential of integrating LLMs with specialized tools to address a wide range of downstream tasks. 
In addition, LLMs have been employed to improve performance on specific chemistry tasks, including reaction prediction~\citep{zhong-etal-2023-reactie, zhong-acl-2024}, drug discovery~\citep{edwards2023synergpt}, and SMILES recognition~\citep{edwards-etal-2021-text2mol}.
Despite growing interest, existing benchmarks often fall short in capturing the unique demands of the chemistry domain, which is inherently knowledge-intensive. 
In contrast to general NLP tasks that frequently involve surface-level reasoning, chemistry requires precise retrieval and synthesis of complex, domain-specific knowledge. 
These characteristics make it a compelling testbed for RAG, where the incorporation of external knowledge sources can substantially enhance LLM reasoning and decision-making.

\section{The \benchmark Benchmark}
\subsection{Evaluation Settings}
The primary goal of this work is to assess RAG systems in a setting that closely mirrors real-world information needs in the chemistry domain while remaining feasible and scalable in practice. To this end, the proposed \benchmark benchmark is designed around four core evaluation scenarios:

\begin{itemize}[leftmargin=*]
    \item \textbf{Zero-Shot Learning:} In real application, demonstrations are hard to find when conducting novel chemistry discovery. Therefore, we do not use any demonstration when evaluating the RAG systems.
    \item \textbf{Open-ended Evaluation:} Most chemistry tasks are open-ended and do not have answer options, including description-guided molecule design, retrosynthesis, and reagent prediction. To better align with chemists' needs, the RAG system should be evaluated in an open-ended setting. In this setting, no answer options will be provided.
    \item \textbf{Multi-Choice Evaluation:} Multiple choice questions are common in LLM-related system evaluation. We adopt a multiple-choice setting to be consistent with previous work, and to make the evaluation more comprehensive. Many open-ended questions can be converted to multiple-choice questions by adding incorrect options.
    \item \textbf{Question-Only Retrieval:} To mimic real-world usage, for multiple-choice questions, only the question is used as the query for RAG.
\end{itemize}

\subsection{Question Datasets}

Our \benchmark contains four datasets that cover a wide range of chemistry tasks, including three multi-choice benchmarks, MMLU-Chem \citep{mmlu}, SciBench \citep{wang2024scibench}, and ChemBench4K \citep{chembench4k}, and one open-ended benchmark, Mol-Instructions \citep{mol-instruct}. MMLU-Chem consists of college chemistry questions collected online. SciBench collects questions from chemistry textbooks. ChemBench4K contains multiple chemical analysis and prediction tasks, but in a multiple-choice fashion. Mol-Instructions is a collection of molecule design, retrosynthesis, and prediction tasks. The statistics of the datasets are shown in Table \ref{tab:benchmark-stats}.

\begin{table}[t]
\small
\centering\setlength{\tabcolsep}{4.0pt}
    \begin{tabular}{c|c|c|c|c}
        \toprule
        \textbf{Dataset} & \textbf{Type} & \textbf{Task} &  \textbf{Size} & \textbf{Avg. Length} \\
        \midrule
        MMLU-Chem & Multi-Choice & Chemistry Understanding & 303 & 31 \\
        \midrule
        SciBench-Chem & Calculation & College-level Examination & 229 & 94 \\
        \midrule
        \multirow{8}{*}{ChemBench4K} & \multirow{8}{*}{Multi-Choice} & Caption2Mol & 100 & \multirow{8}{*}{72} \\
        & & Mol2Caption & 100 &  \\
        & & Name Conversion & 100 &  \\
        & & Product Prediction & 100 &  \\
        & & RetroSynthesis & 100 &  \\
        & & Solvent Prediction & 100 &  \\
        & & Temperature Prediction & 100 &  \\
        & & Yield Prediction & 100 &  \\
        \midrule
        \multirow{6}{*}{Mol-Instructions} & \multirow{6}{*}{Open-Ended} & Desc.-guided Molecule Design & 100 & \multirow{6}{*}{54} \\
        & & Forward Reaction Prediction & 100 &  \\
        & & Molecular Desc. Generation & 100 &  \\
        & & Property Prediction & 100 &  \\
        & & Reagent Prediction & 100 &  \\
        & & RetroSynthesis & 100 &  \\
        \bottomrule
    \end{tabular}
    \caption{Statistics of \benchmark, including question type, task type, data size, and the average length of each question.}
    \label{tab:benchmark-stats}
    \vspace{-3mm}
\end{table}

\paragraph{Metric} For multi-choice questions, we use accuracy as the metric. For open-ended questions, the generated molecule is evaluated by exact match (EM), validity, MACCS FTS, RDK FTS, Morgan FTS, and BLEU. To evaluate the generated text, we use BLEU and ROUGE. For numerical results, we use accuracy with a $5\%$ relative error tolerance. Please refer to Appendix~\ref{app:mol_eval} for more details on molecule evaluation metrics.

\section{The \toolkit}
\toolkit analyzes how RAG systems perform on \benchmark. The \toolkit contains three major components: Corpora, Retrievers, and LLMs.

\paragraph{Corpora} We collect data from six sources: \textsuperscript{1}\href{https://pubchem.ncbi.nlm.nih.gov}{PubChem} for molecule information (English name, SMILES, IUPAC name, weight, mocular formula, and synonyms), \textsuperscript{2}\href{https://pubmed.ncbi.nlm.nih.gov/}{PubMed} for biochemistry abstracts, \textsuperscript{3}\href{https://www.uspto.gov/}{USPTO} for chemical patents information, \textsuperscript{4}\href{https://www.semanticscholar.org/}{Semantic Scholar} for chemistry full-text papers, and \textsuperscript{5}\href{https://openstax.org/}{OpenStax} for chemistry textbooks. The statistics of the corpora are shown in Table \ref{tab:corpora_stats}.

\begin{wraptable}{t}{0.63\textwidth}
\centering\setlength{\tabcolsep}{5.0pt}
\vspace{-3mm}
\small
    \centering
    \begin{tabular}{c|ccc}
        \toprule
        \textbf{Corpus}  & \textbf{\# Snippets} & \textbf{Avg. Length} & \textbf{Domain}\\
        \midrule
        PubChem  & 14.6M & 72 & Chemistry\\
        PubMed & 23.9M & 305 & Biomedicine\\
        USPTO & 143K & 140 & Chemistry\\
        Semantic Scholar  & 32.7M & 403 & Chemistry\\
        OpenStax & 5521 & 273 & Chemistry\\
        Wikipedia & 29.9M & 163 & General\\
        \bottomrule
    \end{tabular}
        \vspace{-3mm}
    \caption{Statistics of corpora in \toolkit. }
    \label{tab:corpora_stats}
\vspace{-3mm}
\end{wraptable}

\paragraph{Retrievers} In \toolkit, we select four representative retrievers for the retrieval process in RAG: BM25 \citep{bm25}, Contriever \citep{contriever}, SPECTER \citep{specter2020cohan}, and e5 \citep{e5}. In addition, we implement Reciprocal Rank Fusion (RRF, \cite{rrf}) to combine the results from different retrievers.

\paragraph{LLMs} We choose a few representative LLMs to be used in \toolkit: Llama-3.1-8B-Instruct, Llama-3.1-70B-Instruct, and Mistral-7B-Instruct-v0.2 for general open-source models, ChemLLM for chemistry open-source model, GPT-3.5-turbo and GPT-4o for closed-source models, Deepseek-R1-Llama-8B and o1 for reasoning models.

\section{Experiment Result}

\subsection{Comparison of Backbone LLMs}
\label{sec:comparison_LLMs}
To systematically study how LLMs perform on chemistry tasks and how the proposed \toolkit affects models, we benchmark various LLMs on \benchmark with the same ChemRAG-Corpora. The top 5 documents retrieved by the RRF retriever are prepended to each question. The results are in Table \ref{tab:main_results}. More implementation details could be found in Appendix~\ref{app:implementation}.

As shown in Table \ref{tab:main_results}, different models behave differently when \toolkit is in use. On average, most models benefit from using \toolkit, Llama-3.1-8B-Instruct gains 25.86\%, Llama-3.1-70B-Instruct gains 24.5\%, Mistral-7B-Instruct gains 36.9\%, GPT-3.5-turbo gains 28.43\%, GPT-4o gains 20.92\%, and o1 gains 16.38\%. The largest improvement often comes from the one in Mol-Instructions and ChemBench4K. Among the backbone LLMs, o1 achieves the highest performance in both baseline and RAG settings. 

Although most models benefit from \toolkit, the performance of ChemLLM decreases slightly ($-12.6\%$) and Deepseek-R1-Llama barely improves. They still gain some performance on certain question datasets. Both ChemLLM and DeepSeek-R1-Llama benefit from RAG on MMLU-Chem ($+14.91\%$ and $+3.59\%$). DeepSeek-R1-Llama also performs slightly better on SciBench and Mol-Instructions with the proposed toolkit ($+0.78$ and $+4.07$). In our experiments, we notice that DeepSeek-R1-Llama-8B does not follow our instructions and generates its answers in various forms, which poses difficulty in parsing its answers and may lead to poor performance in calculation.

We observe that larger models have consistent gains in chemistry-specific benchmarks (SciBench-Chem, ChemBench4K, and Mol-Instructions). This suggests that larger models have a better understanding of the retrieved documents. In MMLU-Chem, most large models (Llama-3.1-70b, GPT-4o, and o1) do not benefit from our toolkit. This may be because MMLU is a common benchmark when evaluating LLMs, and these models are trained on related knowledge. The toolkit may not be able to bring new knowledge to larger models. In SciBench-Chem, many models suffer from using the toolkit, this reflects that these models may not understand the retrieved documents well, since advanced models (Llama-3.1-70b, GPT-4o, and o1) all benefit from the toolkit, and o1 even reaches the highest performance when using the toolkit. In ChemBench4K, similar patterns occur: smaller models have worse results, but larger models gain from the toolkit. In Mol-Instructions, all models gain from the toolkit except ChemLLM.

Since Mol-Instructions contains multiple sub-tasks, and each sub-tasks require multiple metrics, we select description-guided molecule design as a representative to analyze in detail how models perform after using our toolkit. The comparison is shown in Figure \ref{fig:molecule_design}, with more details in the appendix. From Figure \ref{fig:molecule_design}, we observe that with our toolkit, all models improve in all aspects, except ChemLLM.

\begin{table}[t]
\small
    \centering
    \begin{tabular}{c|c|c|c|c|c|c}
        \toprule
        \textbf{LLM} & \textbf{Method} & \textbf{MMLU} & \textbf{SciBench} & \textbf{ChemBench4K} & \textbf{Mol-Instruct.} & \textbf{Avg.}\\
        \midrule
        Llama3.1 & Baseline & 42.90 & 3.30 & 27.25 & 23.99 & 24.36 \\
        (8b) & Ours & 52.15 & 3.56 & 25.88 & 41.05 & 30.66 \\
        \midrule
        Llama3.1 & Baseline & 62.38 & 5.99 & 24.25 & 28.33 & 30.24 \\
        (70b) & Ours & 61.05 & 13.63 & 26.25 & 49.67 & 37.65 \\
        \midrule
        Mistral & Baseline & 45.21 & 2.09 & 12.63 & 4.66 & 16.15 \\
        (7b) & Ours & 42.57 & 0 & 11.13 & 34.73 & 22.11 \\
        \midrule
        ChemLLM & Baseline & 37.62 & 8.72 & 23.5 & 17.74 & 21.90 \\
        (7b) & Ours & 43.23 & 2.03 & 16.75 & 14.56 & 19.14 \\
        \midrule
        Deepseek-r1 & Baseline & 55.44 & 3.09 & 35.38 & 3.75 & 24.42 \\
        -llama(8b) & Ours & 57.43 & 3.87 & 29.13 & 7.82 & 24.56 \\
        \midrule
        \multirow{2}{*}{GPT3.5} & Baseline & 49.17 & 9.66 & 30.5 & 29.00 & 29.58 \\
         & Ours & 52.81 & 8.80 & 44.5 & 45.83 & 37.99 \\
        \midrule
        \multirow{2}{*}{GPT-4o} & Baseline & 74.59 & 4.97 & 59.5 & 28.79 & 41.96 \\
         & Ours & 73.92 & 8.59 & 67.25 & 53.18 & 50.74 \\
        \midrule
        \multirow{2}{*}{o1} & Baseline & 85.81 & 40.82 & 41.63 & 31.55 & 49.95 \\
         & Ours & 85.48 & 43.61 & 58.38 & 45.04 & 58.13 \\
        \bottomrule
    \end{tabular}
    \caption{Benchmark results of different LLMs on \benchmark.}
    \label{tab:main_results}
    \vspace{-3mm}
\end{table}

\begin{figure*}
    \centering
    \includegraphics[width=\textwidth]{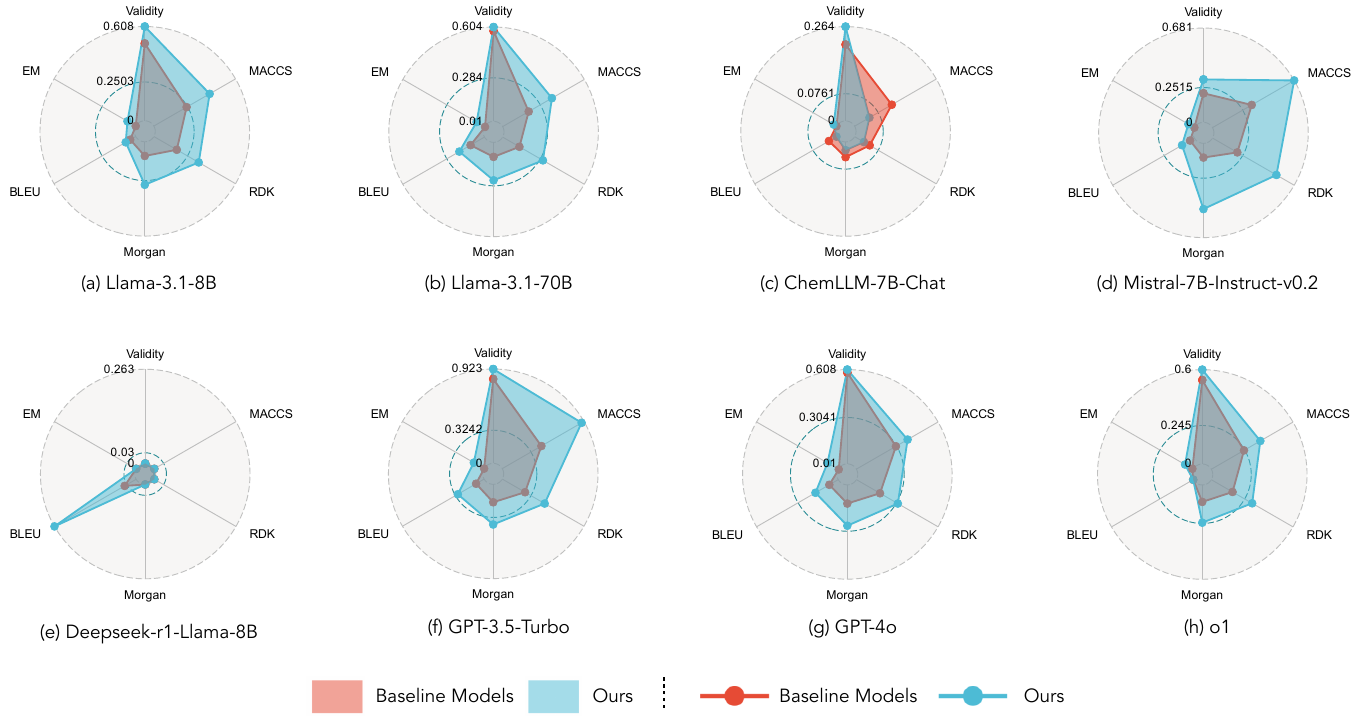}
    \caption{Performance comparison on description-guided molecule design w.r.t evaluation metrics for molecule generation. Ours outperforms the baseline in almost all the scenarios. }
    \label{fig:molecule_design}
    \vspace{-5mm}
\end{figure*}

\begin{table}[t]
\small
    \centering
    \begin{tabular}{c|c|c|c|c|c|c}
        \toprule
        \textbf{Corpus} & \textbf{Retriever} & \textbf{MMLU} & \textbf{SciBench} & \textbf{\shortstack{Chem- \\ Bench4K}} & \textbf{\shortstack{Mol-\\Instructions}} & \textbf{Avg.}\\
        \midrule
        \textit{None} & \textit{None} & 49.17 & 9.66 & 30.5 & 29.00 & 29.58 \\
        \midrule
        \multirow{5}{*}{PubChem} & BM25 & \cellcolor{red!27.9!}47.19 & \cellcolor{green!65.0!}12.98 & \cellcolor{green!25.5!}36.00 & \cellcolor{red!21.0!}27.73 & \cellcolor{green!8.5!}30.98 \\
        & Contriever & \cellcolor{red!13.9!}48.18 & \cellcolor{green!7.05!}10.02 & \cellcolor{green!41.8!}39.50 & \cellcolor{green!1.7!}29.72 & \cellcolor{green!13.8!}31.86 \\
        & SPECTER & \cellcolor{green!10.0!}49.83 & \cellcolor{green!6.27!}9.98 & \cellcolor{green!29.0!}36.75 & \cellcolor{red!36.4!}26.80 & \cellcolor{green!7.6!}30.84 \\
        & e5 & \cellcolor{red!32.5!}46.86 & \cellcolor{red!21.7!}8.61 & \cellcolor{green!46.4!}40.50 & \cellcolor{green!3.9!}30.65 & \cellcolor{green!12.6!}31.66 \\
        & RRF & \cellcolor{red!4.64!}48.84 & \cellcolor{red!12.0!}9.08 & \cellcolor{green!31.9!}37.38 & \cellcolor{green!1.4!}29.58 & \cellcolor{green!9.9!}31.22 \\
        \midrule
        \multirow{5}{*}{PubMed} & BM25 & \cellcolor{red!32.5!}46.86 & \cellcolor{green!46.2!}12.02 & \cellcolor{green!37.7!}38.63 & \cellcolor{red!14.2!}28.14 & \cellcolor{green!11.1!}31.41 \\
        & Contriever & 49.17 & \cellcolor{green!13.9!}10.37 & \cellcolor{green!30.8!}37.13 & \cellcolor{red!24.6!}27.51 & \cellcolor{green!8.9!}31.05 \\
        & SPECTER & \cellcolor{red!27.89!}47.19 & \cellcolor{red!12.0!}9.08 & \cellcolor{green!28.5!}36.63 & \cellcolor{red!21.7!}27.69 & \cellcolor{green!3.5!}30.15 \\
        & e5 & \cellcolor{red!37.1!}46.53 & \cellcolor{green!13.7!}10.36 & \cellcolor{green!42.4!}39.63 & \cellcolor{red!65.0!}25.07 & \cellcolor{green!5.0!}30.40 \\
        & RRF & \cellcolor{red!13.9!}48.18 & \cellcolor{red!14.0!}8.98 & \cellcolor{green!30.8!}37.13 & \cellcolor{red!54.6!}25.70 & \cellcolor{green!2.5!}30.00 \\
        \midrule
        \multirow{5}{*}{USPTO} & BM25 & \cellcolor{green!5.0!}49.50 & \cellcolor{green!37.0!}11.55 & \cellcolor{green!62.7!}44.00 & \cellcolor{green!63.8!}56.17 & \cellcolor{green!65.0!}40.31 \\
        & Contriever & \cellcolor{green!5.0!}49.50 & \cellcolor{green!24.7!}10.92 & \cellcolor{green!54.6!}42.25 & \cellcolor{green!19.7!}37.40 & \cellcolor{green!33.0!}35.02 \\
        & SPECTER & \cellcolor{red!18.6!}47.85 & \cellcolor{green!4.5!}9.44 & \cellcolor{green!30.2!}37.00 & \cellcolor{green!6.4!}31.71 & \cellcolor{red!65.0!}29.00 \\
        & e5 & \cellcolor{red!23.2!}47.52 & \cellcolor{green!27.2!}11.05 & \cellcolor{green!35.4!}38.13 & \cellcolor{green!20.1!}37.55 & \cellcolor{green!24.1!}33.56 \\
        & RRF & 49.17 & \cellcolor{green!20.4!}10.70 & \cellcolor{green!59.8!}43.38 & \cellcolor{green!65.0!}56.68 & \cellcolor{green!63.0!}39.98 \\
        \midrule
        \multirow{5}{*}{\shortstack{Semantic \\ Scholar}} & BM25 & \cellcolor{red!51.1!}45.54 & \cellcolor{red!51.2!}7.18 & \cellcolor{green!31.3!}37.25 & \cellcolor{green!1.7!}29.72 & \cellcolor{green!2.1!}29.92 \\
        & Contriever & \cellcolor{red!18.6!}47.85 & \cellcolor{green!58.5!}12.65 & \cellcolor{green!38.9!}38.88 & \cellcolor{green!6.4!}31.73 & \cellcolor{green!19.4!}32.78 \\
        & SPECTER & 49.17 & \cellcolor{green!15.5!}10.45 & \cellcolor{green!30.2!}37.00 & \cellcolor{red!42.3!}26.44 & \cellcolor{green!7.2!}30.77 \\
        & e5 & \cellcolor{red!55.7!}45.21 & \cellcolor{green!22.0!}10.78 & \cellcolor{green!38.3!}38.75 & \cellcolor{green!5.9!}31.52 & \cellcolor{green!12.1!}31.57 \\
        & RRF & \cellcolor{red!65.0!}44.55 & \cellcolor{red!15.5!}8.91 & \cellcolor{green!40.1!}39.13 & \cellcolor{green!6.5!}31.76 & \cellcolor{green!9.1!}31.09 \\
        \midrule
        \multirow{5}{*}{OpenStax} & BM25 & \cellcolor{green!20.2!}50.17 & \cellcolor{green!7.4!}10.04 & \cellcolor{green!34.3!}37.88 & \cellcolor{red!10.9!}28.34 & \cellcolor{green!12.3!}31.61 \\
        & Contriever & \cellcolor{green!5.0!}49.50 & \cellcolor{green!58.7!}12.66 & \cellcolor{green!27.3!}36.38 & \cellcolor{red!17.4!}27.95 & \cellcolor{green!12.4!}31.62 \\
        & SPECTER & \cellcolor{green!20.2!}50.50 & \cellcolor{green!43.5!}11.88 & \cellcolor{green!30.8!}37.13 & \cellcolor{red!13.2!}28.20 & \cellcolor{green!14.2!}31.93 \\
        & e5 &\cellcolor{green!10.0!} 49.83 & \cellcolor{green!37.4!}11.57 & \cellcolor{green!37.1!}38.5 & \cellcolor{green!2.3!}29.96 & \cellcolor{green!17.5!}32.47 \\
        & RRF &\cellcolor{green!50.2!} 52.48 & \cellcolor{green!37.0!}11.55 & \cellcolor{green!31.3!}37.25 & \cellcolor{green!0.8!}29.35 & \cellcolor{green!18.7!}32.66 \\
        \midrule
        \multirow{5}{*}{Wiki} & BM25 & 49.17 & \cellcolor{red!33.0!}8.06 & \cellcolor{green!38.3!}38.75 & \cellcolor{red!22.0!}27.67 & \cellcolor{green!8.1!}30.91 \\
        & Contriever & \cellcolor{red!4.6!}48.84 & \cellcolor{green!13.1!}10.33 & \cellcolor{green!31.3!}37.25 & \cellcolor{green!0.3!}29.14 & \cellcolor{green!11.0!}31.39 \\
        & SPECTER & \cellcolor{red!23.2!}47.52 & \cellcolor{red!9.3!}9.21 & \cellcolor{green!40.6!}39.25 & \cellcolor{red!29.4!}27.22 & \cellcolor{green!7.4!}30.8 \\
        & e5 & \cellcolor{green!25.2!}50.83 & \cellcolor{red!15.1!}8.93 & \cellcolor{green!30.8!}37.13 & \cellcolor{green!1.3!}29.54 & \cellcolor{green!12.3!}31.61 \\
        & RRF & \cellcolor{green!15.2!}50.17 & \cellcolor{green!20.4!}10.70 & \cellcolor{green!34.8!}38.00 & \cellcolor{red!22.2!}27.66 & \cellcolor{green!12.4!}31.63 \\
        \midrule
        \multirow{5}{*}{\shortstack{Chem \\ -RAG \\ Corpus}} & BM25 & \cellcolor{green!10.0!}49.83 & \cellcolor{red!65.0!}6.51 & \cellcolor{green!35.4!}38.13 & \cellcolor{green!14.0!}34.99 & \cellcolor{green!16.9!}32.37 \\
        & Contriever & \cellcolor{green!65.0!}53.46 & \cellcolor{green!57.2!}12.58 & \cellcolor{green!56.3!}42.63 & \cellcolor{green!30.7!}42.08 & \cellcolor{green!49.1!}37.69 \\
        & SPECTER & \cellcolor{red!13.9!}48.18 & \cellcolor{red!22.5!}8.57 & \cellcolor{green!51.7!}41.63 & \cellcolor{green!7.9!}32.35 & \cellcolor{green!18.8!}32.69 \\
        & e5 & \cellcolor{red!27.9!}47.19 & \cellcolor{red!43.3!}7.56 & \cellcolor{green!30.8!}37.13 & \cellcolor{green!31.1!}42.24 & \cellcolor{green!23.9!}33.53 \\
        & RRF & \cellcolor{green!55.2!}52.81 & \cellcolor{red!17.7!}8.80 & \cellcolor{green!65.0!}44.5 & \cellcolor{green!39.5!}45.83 & \cellcolor{green!50.9!}37.99 \\
        \bottomrule
    \end{tabular}
    \caption{Experiment results of various retrievers and corpora on \benchmark. Compared with the baseline (first row), the intensity of the shade represents the magnitude of the \colorbox{red!50}{decreases} and \colorbox{green!50}{increases}.}
    \label{tab:retriever_analysis}
    \vspace{-3mm}
\end{table}

\subsection{Comparison of Retrievers and Corpora}
To understand the effect of each component in \toolkit, we benchmark different retrievers with different corpora on \benchmark. The experiments are conducted with GPT-3.5-turbo since it is one of the models that benefit most from our toolkit, and it is also efficient and inexpensive for inference. The results are in Table \ref{tab:retriever_analysis}.

\paragraph{Comparison between Corpora} From Table \ref{tab:retriever_analysis}, we observe that the performance of a RAG system is correlated to the selected corpus. The model performs the best with OpenStax (textbook) on MMLU-Chem and SciBench-Chem, but OpenStax barely has benefit for Mol-Instructions. USPTO helps the model to achieve its best on ChemBench4K and Mol-Instructions, but it provides little benefit on MMLU-Chem and SciBench-Chem. When using the combined \chemrag Corpus, the model achieves the best on MMLU-Chem and ChemBench4K, surpassing leveraging only one corpus, which demonstrates the significance of combining multiple corpora. The \chemrag Corpus also helps the model to perform better on Mol-Instructions, only not as good as USPTO. Our corpus is also beneficial for SciBench when using Contriever as the retriever. 

\paragraph{Comparison between Retrievers} The Retriever plays another critical role as it decides how the documents rank. From our experiments shown in Table \ref{tab:retriever_analysis}, all retrievers have their best performance on a specific corpus and task. BM25 shows a very strong performance when using USPTO on Mol-Instructions, and using PubChem on SciBench-Chem. Contriever outperforms other retrievers when incorporating \chemrag Corpus on MMLU-Chem, it also works well with PubChem and the \chemrag Corpus. SPECTER and e5 have mixed performances but still can excel in certain corpora. For instance, SPECTER improves the most when using Wikipedia on ChemBench4K. e5 surpasses other retrievers on Mol-Instructions when using Wikipedia. The RRF retriever, combining the results of the four retrievers, usually improves the performance, even though it might not be the best, and sometimes results in the best performance. For instance, RRF helps the model achieve the best on MMLU-Chem and ChemBench4K.

\section{Discussion and Analyses}
\begin{figure*}[t]
    \centering
    \includegraphics[width=0.95\textwidth]{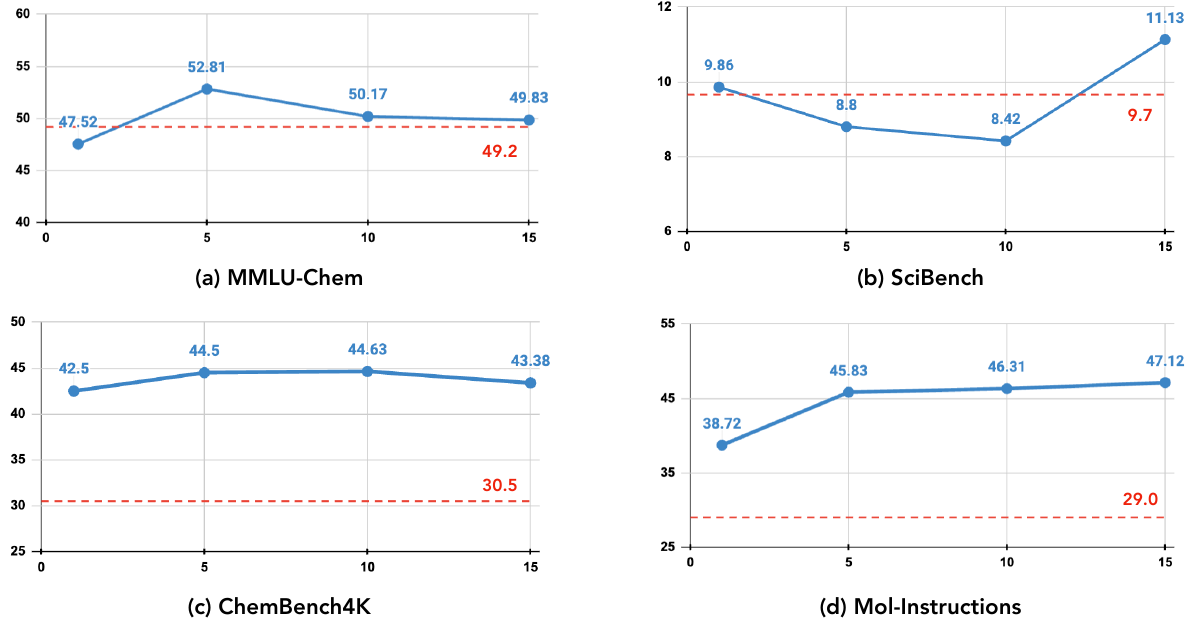}
    \vspace{-3mm}
    \caption{Performance comparison on different numbers of retrieved documents. The \textcolor{red}{red} dotted line represents the baseline. The experiments are conducted on GPT-3.5-turbo.}
    \label{fig:scaling_study}
    \vspace{-5mm}
\end{figure*}
\subsection{Performance Scaling}
The number of retrieved documents $k$ is an important factor in RAG systems. When $k$ is too small, RAG systems may lack critical information; on the other hand, when $k$ is too large, RAG systems may suffer from too much irrelevant information. To better understand how this factor affects RAG systems, we conduct experiments on $k=1, 5, 10, 15$. The results are shown in Figure \ref{fig:scaling_study}.
We can see that the phenomenon where performance first increases and then decreases as $k$ increases is clearly observed in MMLU-Chem and ChemBench4K. In Mol-Instructions, though the performances increase when $k$ increases, the difference is very small when $k\geq 5$. The performance only increases $1.29$ when $k$ increase from $5$ to $15$. In SciBench-Chem, the performance first decreases but then increases. This suggests that a better retriever is needed or a reranker should be used. In our opinion, a better retriever should be developed since current retrievers only consider semantic similarity, however, semantic similarity may not be sufficient in reasoning tasks like SciBench.
Overall, $k=5$ is a good choice since it provides sufficient information in most cases.

\subsection{Proportion in the \chemrag Corpus}
\begin{wrapfigure}{r}{0.6\textwidth}
\vspace{-5mm}
    \centering
    \includegraphics[width=0.75\textwidth]{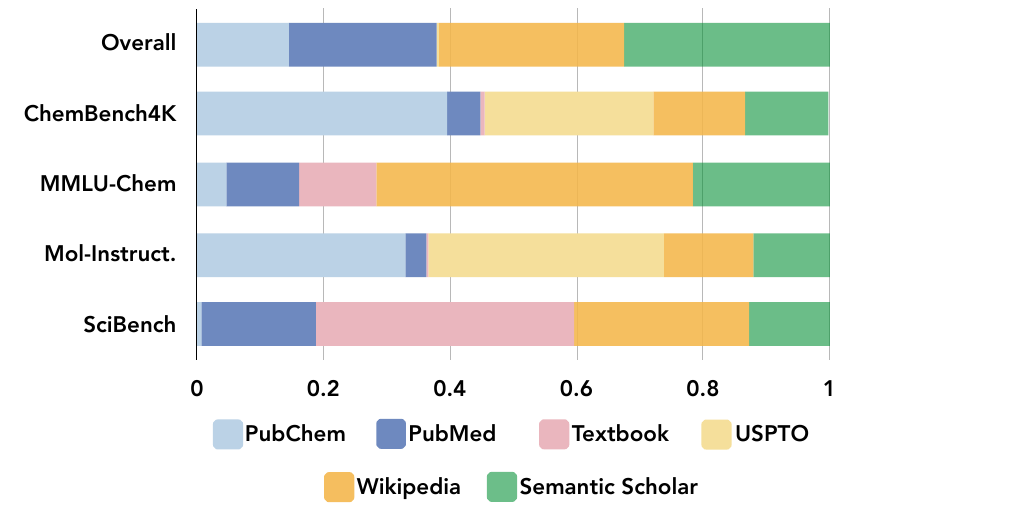}
    \vspace{-3mm}
    \caption{The overall corpus composition of \chemrag corpora
    and the actually retrieved proportion in different tasks.}
    \label{fig:corpus_proportion}
    \vspace{-3mm}
\end{wrapfigure}

We investigate the proportion of different sources used across various tasks. Figure \ref{fig:corpus_proportion} shows the proportions of six sources in \chemrag-Corpus, and the actual proportions in the top 50 retrieved chunks in \benchmark. A task-specific pattern of proportion is observed. OpenStax has a larger proportion in SciBench and a relatively large proportion in MMLU-Chem. This is natural since the questions in SciBench and MMLU are derived from academic settings. PubChem has the largest proportion in both ChemBench4K and Mol-Instructions, which can be explained by the fact that these two tasks focus on molecule-related questions.

\subsection{Retrievers in Chemistry}
In our observation from Table \ref{tab:retriever_analysis} and Figure \ref{fig:scaling_study}, we believe that a better retriever is needed for retrieving documents for chemistry downstream tasks. In Table \ref{tab:retriever_analysis}, the model always performs better with USPTO and OpenStax (textbook) corpora, but it performs worse on the combined corpus, which suggests the retriever ranks the helpful snippets to a lower place. This is also validated by the sudden rise in Figure \ref{fig:scaling_study} (b).

In addition, chemistry retrieval faces a ``multi-modality`` issue. One chemical compound may have multiple representations, including SMILES strings, IUPAC names, and English names, and each of them has variants. It is likely that a SMILES string is mentioned in a question, but the related information is in a chemistry paper, which usually uses English names instead of SMILES. Unfortunately, current retrievers cannot solve this problem.

Finally, current retrievers only consider semantic similarities, but chemistry tasks require more. For example, when predicting the yield of a reaction, one may want to search for the yield of a similar reaction type instead of searching for a match with chemical compounds.

\subsection{Practical Recommendations}
Based on our experiments, we provide some practical recommendations:
\begin{itemize}[leftmargin=*]

    \item\textbf{Corpus Selection}
The proposed \chemrag-Corpus is a good start and is likely to outperform using only one corpus source. This is confirmed in Table \ref{tab:retriever_analysis}, MMLU-Chem and ChemBench4K in particular. When working on molecule-related tasks, one may want to try USPTO since it reaches high performance in both ChemBench4K and Mol-Instructions. As for questions in school, OpenStax (textbook) may be preferred, but the performance is still lower than using \chemrag-Corpus in MMLU-Chem, illustrated in Table \ref{tab:retriever_analysis}.

\item\textbf{Retriever Selection}
Contriever is the most stable retriever in the four individual retrievers, but its performance still fluctuates across tasks and corpora. The proposed RRF retriever is recommended since it usually performs close to the best individual retriever and sometimes outperforms them.

\item\textbf{LLM Selection}
o1 is the best model for all the tasks. Considering the cost and inference speed, GPT-3.5-turbo and GPT-4o are good options. For open-source models, Llama-3.1-8B-Instruct is preferred since it achieves the second among the five open-source models and performs similar to the best model, Llama-3.1-70B-Instruct. Llama-3.1-70B only performs 24\% better, but with 775\% more parameters and much higher computation cost. 
\end{itemize}

\section{Conclusion}
We propose \benchmark and \toolkit to systematically evaluate RAG systems in chemistry. Based on our extensive experiments, we provide some novel findings, practical recommendations, and future directions for the community to better leverage RAG systems in chemistry in the real-world.

\section{Acknowledgments}
Research was supported in part by US DARPA INCAS Program No. HR0011-21-C0165 and BRIES Program No. HR0011-24-3-0325, National Science Foundation IIS-19-56151, the Molecule Maker Lab Institute: An AI Research Institutes program supported by NSF under Award No. 2019897, the Institute for Geospatial Understanding through an Integrative Discovery Environment (I-GUIDE) by NSF under Award No. 2118329, and Apple PhD Fellowship. This research was also supported in part by the Division of Intramural Research (DIR), National Library of Medicine (NLM), National Institutes of Health (NIH). Any opinions, findings, and conclusions or recommendations expressed herein are those of the authors and do not necessarily represent the views, either expressed or implied, of DARPA or the U.S. Government. This research has used the Delta/DeltaAI advanced computing and data resource, a joint effort of the University of Illinois Urbana-Champaign and its National Center for Supercomputing Applications, supported by the National Science Foundation (award OAC 2320345) and the State of Illinois. Access to DeltaAI was provided in part by the Illinois Computes project which is supported by the University of Illinois Urbana-Champaign and through allocation \#250851 from the Advanced Cyberinfrastructure Coordination Ecosystem: Services \& Support (ACCESS) program, which is supported by National Science Foundation grants \#2138259, \#2138286, \#2138307, \#2137603, and \#2138296. The views and conclusions contained in this paper are those of the authors and should not be interpreted as representing any funding agencies.

\newpage
\bibliography{colm2025_conference}
\bibliographystyle{colm2025_conference}

\appendix
\section{Evaluation Metrics for Molecules}
\label{app:mol_eval}
To assess the quality of generated molecules, we first employ general text-based generation metrics such as BLEU~\citep{DBLP:conf/acl/PapineniRWZ02} and ROUGE~\citep{lin2004rouge}, which compare generated outputs against reference answers.

For molecular generation, we begin by verifying the validity of generated molecules using RDKit~\citep{landrum2013rdkit} and then compute their exact match with reference solutions. However, a single textual description can correspond to multiple molecular structures, making exact matching a limited evaluation criterion. Moreover, expecting an LLM—even one fine-tuned with LoRA on specific instructions—to consistently generate outputs that perfectly match reference molecules is often unrealistic.

To address these challenges and provide a more comprehensive evaluation, we incorporate molecular similarity metrics, including similarity scores based on RDKit, MACCS, and Morgan fingerprints~\citep{tanimoto1958elementary,schneider2015get,durant2002reoptimization}, alongside Levenshtein~\citep{yujian2007normalized} and BLEU scores.

For tasks that need to compare numbers, following previous work~\citep{wang2024scibench}, we compare the generated output with the ground truth, allowing a 5\% relative error. This makes sure that the score is within $0$ and $1$, making it more suitable for combining with other scores. These results can be found in Appendix \ref{app:experiment_results}.

\section{Implementation Details}\label{app:implementation}
Since DeepSeek-R1-Llama and o1 are reasoning models, following their guidelines, we set 0.6 and 1 as their temperatures respectively. The temperatures for other models are set to 0 for reproducibility. 
For each experiment, we run three rounds and report the mean for DeepSeek-R1-Llama and o1 models.
We run one round for other models. The number of max generation tokens is set to 10,000 for DeepSeek-R1-Llama and o1 since their reasoning requires more tokens, and the numbers for other models are set to 512.

\section{Detailed Experiment Results}\label{app:experiment_results}
\subsection{SciBench-Chemistry}
Table \ref{tab:scibench_result} shows the performances of different models on SciBench-Chemistry tasks.

\begin{table}[t]
\small
    \centering
    \begin{tabular}{c|c|c|c|c|c|c}
        \toprule
        \textbf{LLM} & \textbf{Method} &  \multicolumn{4}{|c|}{\textbf{SciBench}}  & \textbf{Avg.}\\
        \midrule
        \multicolumn{2}{c|}{} & atkins & chemmec & matter & quan & \\
        \midrule
        Llama3.1 & Baseline & 0 & 10.26 & 0 & 2.94 & 3.30\\
        (8b) & Ours & 3.74 & 2.56 & 2.04 & 5.88 & 3.56 \\
        \midrule
        Llama3.1 & Baseline & 2.56 & 5.99 & 0 & 17.65 & 5.99 \\
        (70b) & Ours & 6.54 & 15.38 & 6.12 & 26.47 & 13.63 \\
        \midrule
        Mistral & Baseline & 3.74 & 2.56 & 2.04 & 0 & 2.09 \\
        (7b) & Ours & 0 & 0 & 0 & 0 & 0 \\
        \midrule
        ChemLLM & Baseline & 11.21 & 12.82 & 2.04 & 8.82 & 8.72 \\
        (7b) & Ours & 0.93 & 5.13 & 2.04 & 0 & 2.03 \\
        \midrule
        Deepseek-r1 & Baseline & 4.67 & 7.69 & 0 & 0 & 3.09 \\
        -llama(8b) & Ours & 2.80 & 7.69 & 2.04 & 2.94 & 3.87 \\
        \midrule
        \multirow{2}{*}{GPT3.5} & Baseline & 5.61 & 23.08 & 4.08 & 5.88 & 9.66 \\
         & Ours & 5.61 & 20.51 & 6.12 & 2.94 & 8.80 \\
        \midrule
        \multirow{2}{*}{GPT-4o} & Baseline & 3.74 & 10.26 & 0 & 5.88 & 4.97 \\
         & Ours & 10.28 & 10.26 & 2.04 & 11.76 & 8.59 \\
        \midrule
        \multirow{2}{*}{o1} & Baseline & 38.32 & 46.15 & 34.69 & 44.12 & 40.82 \\
         & Ours & 44.86 & 48.72 & 36.73 & 44.12 & 43.61 \\
        \bottomrule
    \end{tabular}
    \caption{Detailed benchmark results of different LLMs on SciBench-Chemistry. The accuracy is computed by comparing the generated answer with the ground truth, allowing a 5\% relative error.}
    \label{tab:scibench_result}
    \vspace{-3mm}
\end{table}

\subsection{ChemBench4K}
Table \ref{tab:chembench4k_result1} and Table \ref{tab:chembench4k_result2} demonstrate the performances of different models on ChemBench4K tasks.

\begin{table}[t]
\small
    \centering
    \begin{tabular}{c|c|c|c|c|c}
        \toprule
        \textbf{LLM} & \textbf{Method} &  \multicolumn{4}{|c}{\textbf{ChemBench4K}} \\
        \midrule
        \multicolumn{2}{c|}{} & \multirow{2}{*}{Caption2Mol} & \multirow{2}{*}{Mol2Caption} & Name & Product \\
        \multicolumn{2}{c|}{} &  &  & Conversion & Prediction \\
        \midrule
        Llama3.1 & Baseline & 0 & 88 & 57 & 13 \\
        (8b) & Ours & 7 & 70 & 59 & 15 \\
        \midrule
        Llama3.1 & Baseline & 3 & 86 & 68 & 0 \\
        (70b) & Ours & 5 & 87 & 64 & 11\\
        \midrule
        Mistral & Baseline & 3 & 26 & 40 & 15 \\
        (7b) & Ours & 7 & 19 & 33 & 11\\
        \midrule
        ChemLLM & Baseline & 24 & 46 & 48 & 2\\
        (7b) & Ours & 21 & 33 & 38 & 0 \\
        \midrule
        Deepseek-r1 & Baseline & 15 & 81 & 70 & 19 \\
        -llama(8b) & Ours & 18 & 68 & 65 & 12 \\
        \midrule
        \multirow{2}{*}{GPT3.5} & Baseline & 20 & 89 & 48 & 17 \\
         & Ours & 36 & 87 & 60 & 39 \\
        \midrule
        \multirow{2}{*}{GPT-4o} & Baseline & 41 & 98 & 79 & 93\\
         & Ours & 61 & 98 & 81 & 83\\
        \midrule
        \multirow{2}{*}{o1} & Baseline & 6 & 99 & 76 & 26 \\
         & Ours & 27 & 96 & 80 & 59\\
        \bottomrule
    \end{tabular}
    \caption{Detailed benchmark results of different LLMs on ChemBench4K, Part 1.}
    \label{tab:chembench4k_result1}
    \vspace{-3mm}
\end{table}

\begin{table}[t]
\small
    \centering
    \begin{tabular}{c|c|c|c|c|c}
        \toprule
        \textbf{LLM} & \textbf{Method} &  \multicolumn{4}{|c}{\textbf{ChemBench4K}} \\
        \midrule
        \multicolumn{2}{c|}{} & \multirow{2}{*}{Retrosynthesis} & Solvent & Temp. & Yield \\
        \multicolumn{2}{c|}{} &  & Prediction & Prediction & Prediction \\
        \midrule
        Llama3.1 & Baseline & 0 & 21 & 17 & 22\\
        (8b) & Ours & 6 & 20 & 15 & 15 \\
        \midrule
        Llama3.1 & Baseline & 0 & 24 & 9 & 4 \\
        (70b) & Ours & 2 & 22 & 1 & 18 \\
        \midrule
        Mistral & Baseline & 0 & 2 & 9 &  6\\
        (7b) & Ours & 2 & 8 & 0  & 9\\
        \midrule
        ChemLLM & Baseline & 1 & 25 & 21 & 21\\
        (7b) & Ours & 0 & 28 & 4  &10 \\
        \midrule
        Deepseek-r1 & Baseline & 14 & 36 & 31 & 17 \\
        -llama(8b) & Ours & 3 & 33 & 17 & 17 \\
        \midrule
        \multirow{2}{*}{GPT3.5} & Baseline & 4 & 23 & 18 & 25 \\
         & Ours & 26 & 41 &  28 & 39\\
        \midrule
        \multirow{2}{*}{GPT-4o} & Baseline & 54 & 35 & 33 & 43\\
         & Ours & 76 & 49 & 43 & 47\\
        \midrule
        \multirow{2}{*}{o1} & Baseline & 5 & 42 & 48 & 31 \\
         & Ours & 50 & 50 & 63 & 42\\
        \bottomrule
    \end{tabular}
    \caption{Detailed benchmark results of different LLMs on ChemBench4K, Part 2.}
    \label{tab:chembench4k_result2}
    \vspace{-3mm}
\end{table}

\subsection{Mol-Instructions}
Table \ref{tab:mol-instruct_result1}, Table \ref{tab:mol-instruct_result2}, Table \ref{tab:mol-instruct_result3}, and Table \ref{tab:mol-instruct_result4} demonstrate some results in Mol-Instructions.

\begin{table}[t]
\small
    \centering
    \begin{tabular}{c|c|c|c|c|c|c|c}
        \toprule
        \textbf{LLM} & \textbf{Method} &  \multicolumn{6}{|c}{\textbf{Description-Guided Molecule Deisgn}} \\
        \midrule
        \multicolumn{2}{c|}{} & \multirow{2}{*}{EM$\uparrow$} & \multirow{2}{*}{Validity$\uparrow$} & MACCS & RDK & Morgan & \multirow{2}{*}{BLEU$\uparrow$} \\
        \multicolumn{2}{c|}{} & & & FTS$\uparrow$ & FTS$\uparrow$ & FTS$\uparrow$ & \\
        \midrule
        Llama3.1 & Baseline & 0	& 73 & 35.71 & 25.01 & 13.37 & 6.02 \\
        (8b) & Ours & 9 & 89 & 60.78 & 48.85 & 40.64 & 10.92 \\
        \midrule
        Llama3.1 & Baseline & 1 & 95 & 32.61 & 21.67 & 16.72 & 18.34 \\
        (70b) & Ours & 11 & 99 & 60.35 & 49.65 & 40.89 & 31.56 \\
        \midrule
        Mistral & Baseline & 0 & 21 & 32.74 & 20.66 & 10.34 & 3.61 \\
        (7b) & Ours & 5 & 31 & 68.14 & 53.26 & 47.52 & 10.35 \\
        \midrule
        ChemLLM & Baseline & 0 & 47 & 26.40  & 10.70  & 9.41 & 5.41 \\
        (7b) & Ours & 2 & 58 & 10.48 & 6.59 & 5.10 & 0 \\
        \midrule
        Deepseek-r1 & Baseline & 0 & 0 & 0 & 0 & 0 & 3.70 \\
        -llama(8b) & Ours & 0 & 0 & 0 & 0 & 0 & 26.25 \\
        \midrule
        \multirow{2}{*}{GPT3.5} & Baseline & 0 & 85 & 45.53 & 26.48 & 18.08 & 9.41 \\
         & Ours & 12 & 95 & 92.28 & 49.35 & 40.45 & 30.56 \\
        \midrule
        \multirow{2}{*}{GPT-4o} & Baseline & 1 & 93 & 47.33 & 28.88 & 20.32 & 11.88 \\
         & Ours & 14 & 96 & 60.84 & 49.47 & 42.45 & 27.92 \\
        \midrule
        \multirow{2}{*}{o1} & Baseline & 1 & 89 & 40.12 & 25.72 & 17.55 & - \\
         & Ours & 12 & 97 & 57.59 & 46.01 & 39.78 & - \\
        \bottomrule
    \end{tabular}
    \caption{Detailed benchmark results of different LLMs on Mol-Instructions -- Description-guided molecule design.}
    \label{tab:mol-instruct_result1}
    \vspace{-3mm}
\end{table}

\begin{table}[t]
\small
    \centering
    \begin{tabular}{c|c|c|c|c|c|c|c}
        \toprule
        \textbf{LLM} & \textbf{Method} &  \multicolumn{6}{|c}{\textbf{Forward Reaction Prediction}} \\
        \midrule
        \multicolumn{2}{c|}{} & \multirow{2}{*}{EM$\uparrow$} & \multirow{2}{*}{Validity$\uparrow$} & MACCS & RDK & Morgan & \multirow{2}{*}{BLEU$\uparrow$} \\
        \multicolumn{2}{c|}{} & & & FTS$\uparrow$ & FTS$\uparrow$ & FTS$\uparrow$ & \\
        \midrule
        Llama3.1 & Baseline & 0 & 38 & 59 & 60.24 & 43.19 & 6.63\\
        (8b) & Ours & 17 & 92 & 63.29 & 53.61 & 45.46 & 29.50 \\
        \midrule
        Llama3.1 & Baseline & 0 &68  & 63.74 & 60.68 & 44.94 & 17.47 \\
        (70b) & Ours & 22 &91  & 72.70 & 62.74 & 57.14 & 43.89 \\
        \midrule
        Mistral & Baseline & 0 & 0 & 0 & 0 & 0 & 2 \\
        (7b) & Ours & 3 & 29 & 76.68 & 77.54 & 63.33 & 11.31 \\
        \midrule
        ChemLLM & Baseline & 0 & 29 & 45.88 & 34.61 & 28.15 & 3.18 \\
        (7b) & Ours & 0 & 57 & 22.55 & 17.55 & 12.68 & 0 \\
        \midrule
        Deepseek-r1 & Baseline & 0 & 33 & 1.56 & 0.39 & 0.65 & 12.98 \\
        -llama(8b) & Ours & 0 & 59 & 0 & 0 & 0 & 1.77 \\
        \midrule
        \multirow{2}{*}{GPT3.5} & Baseline & 0 & 57 & 58.37 & 52.03 & 40.63 & 23.43 \\
         & Ours & 16 & 96 & 72.11 & 67.80 & 56.21 & 39.43 \\
        \midrule
        \multirow{2}{*}{GPT-4o} & Baseline & 2 & 96 & 66.35 & 62.6 & 50.84 & 50.7 \\
         & Ours & 26 & 89 & 78.31 & 73.88 & 68.3 & 61.44 \\
        \midrule
        \multirow{2}{*}{o1} & Baseline & 13 &  87&  81.95& 78.95 & 72.06 & - \\
         & Ours & 30 & 90 & 87.35 & 82.89 & 78.92 & -  \\
        \bottomrule
    \end{tabular}
    \caption{Detailed benchmark results of different LLMs on Mol-Instruction -- Forward Reaction Prediction.}
    \label{tab:mol-instruct_result2}
    \vspace{-3mm}
\end{table}

\begin{table}[t]
\small
    \centering
    \begin{tabular}{c|c|c|c}
        \toprule
        \textbf{LLM} & \textbf{Method} &  \multicolumn{2}{|c}{\textbf{Molecule Description Generation}} \\
        \midrule
        \multicolumn{2}{c|}{} & BLEU & Rouge-L \\
        \midrule
        Llama3.1 & Baseline & 0 & 8.98  \\
        (8b) & Ours &  8.24 & 32.79   \\
        \midrule
        Llama3.1 & Baseline & 0.83 &  15.25  \\
        (70b) & Ours &  4.30 & 27.6   \\
        \midrule
        Mistral & Baseline & 0.63 &  18.64 \\
        (7b) & Ours & 4.48 & 32.09   \\
        \midrule
        ChemLLM & Baseline & 5.33 & 34.04   \\
        (7b) & Ours & 0 &  0  \\
        \midrule
        Deepseek-r1 & Baseline & 0 & 0  \\
        -llama(8b) & Ours & 0 &  0  \\
        \midrule
        \multirow{2}{*}{GPT3.5} & Baseline & 3.18 & 20.58   \\
         & Ours & 3.75 & 21.51   \\
        \midrule
        \multirow{2}{*}{GPT-4o} & Baseline & 1.23 & 18.25   \\
         & Ours & 2.98 & 30.06   \\
        \midrule
        \multirow{2}{*}{o1} & Baseline & 0 &  0  \\
         & Ours & 1.02 & 14.44   \\
        \bottomrule
    \end{tabular}
    \caption{Detailed benchmark results of different LLMs on Mol-Instructions -- Moleclue Description Generation.}
    \label{tab:mol-instruct_result3}
    \vspace{-3mm}
\end{table}

\begin{table}[t]
\small
    \centering
    \begin{tabular}{c|c|c}
        \toprule
        \textbf{LLM} & \textbf{Method} &  \textbf{Property Prediction} \\
        \midrule
        \multicolumn{2}{c|}{} & Accuracy \\
        \midrule
        Llama3.1 & Baseline & 0.14  \\
        (8b) & Ours &  0   \\
        \midrule
        Llama3.1 & Baseline &  0 \\
        (70b) & Ours &  0   \\
        \midrule
        Mistral & Baseline & 0 \\
        (7b) & Ours &  0  \\
        \midrule
        ChemLLM & Baseline &  60  \\
        (7b) & Ours &  15 \\
        \midrule
        Deepseek-r1 & Baseline &  1 \\
        -llama(8b) & Ours &  1 \\
        \midrule
        \multirow{2}{*}{GPT3.5} & Baseline & 18   \\
         & Ours &  1  \\
        \midrule
        \multirow{2}{*}{GPT-4o} & Baseline &  2 \\
         & Ours &   0 \\
        \midrule
        \multirow{2}{*}{o1} & Baseline &  0 \\
         & Ours &  0  \\
        \bottomrule
    \end{tabular}
    \caption{Detailed benchmark results of different LLMs on Mol-Instructions -- Property Prediction. The accuracy is computed by comparing the generated answer with the ground truth, allowing a 5\% relative error.}
    \label{tab:mol-instruct_result4}
    \vspace{-3mm}
\end{table}

\section{Prompt}
The prompts used in our experiments can be found in Table \ref{tab:prompt}, \ref{tab:prompt1}, \ref{tab:prompt2}, \ref{tab:prompt3}, \ref{tab:prompt4}, \ref{tab:prompt5}, \ref{tab:prompt6}, \ref{tab:prompt7}.

\begin{table*}[t]
    \centering
    
    \caption{Baseline prompt template for general open-ended questions.}
    \label{tab:prompt}    
    \begin{minipage}{0.95\columnwidth}
        \centering
        \begin{tcolorbox}[title=Open-ended Baseline Prompt]
            Answer the question directly.
            
            Only give me the answer and do not output any other words.

            \textit{Question: \{ Instruction \}}
            
            \textit{Answer:}
        \end{tcolorbox}
    \end{minipage}
    
\end{table*}

\begin{table*}[t]
    \centering
    
    \caption{Multi-choice baseline prompt template for general open-ended questions.}
    \label{tab:prompt1}    
    \begin{minipage}{0.95\columnwidth}
        \centering
        \begin{tcolorbox}[title=Multi-choice Baseline Prompt]
            Answer the question directly.
            
            Only give me the answer and do not output any other words.

            \textit{Question: \{ Instruction \}}
            
            \textit{Choices: \{ Choices \}}

            Make prediction from the given choices.
            
            \textit{Answer:}
        \end{tcolorbox}
    \end{minipage}
    
\end{table*}

\begin{table*}[t]
    \centering
    
    \caption{Numerical baseline prompt template for general open-ended questions.}
    \label{tab:prompt2}    
    \begin{minipage}{0.95\columnwidth}
        \centering
        \begin{tcolorbox}[title=Numerical Baseline Prompt]
            Answer the question directly.

            \textbf{Conclude the answer by stating ``The answer is therefore [ANSWER]``}
            
            Only give me the answer and do not output any other words.

            \textit{Question: \{ Instruction \}}
            
            \textit{Answer:}
        \end{tcolorbox}
    \end{minipage}
    
\end{table*}

\begin{table*}[t]
    \centering
    
    \caption{Generation baseline prompt template for general open-ended questions.}
    \label{tab:prompt3}    
    \begin{minipage}{0.95\columnwidth}
        \centering
        \begin{tcolorbox}[title=Generation Baseline Prompt]
            Answer the question directly.

            Your answer should be surrounded by [ANSWER] and [/ANSWER]. When generating a molecule, please generate a valid SMILES string.
            
            Only give me the answer and do not output any other words.

            \textit{Question: \{ Instruction \}}
            
            \textit{Answer:}
        \end{tcolorbox}
    \end{minipage}
    
\end{table*}

\begin{table*}[t]
    \centering
    
    \caption{RAG prompt template for general open-ended questions.}
    \label{tab:prompt4}    
    \begin{minipage}{0.95\columnwidth}
        \centering
        \begin{tcolorbox}[title=Open-ended RAG Prompt]
            Answer the question based on the given document.
            
            Only give me the answer and do not output any other words.

            The following are given documents.

            \{ reference \}

            \textit{Question: \{ Instruction \}}
            
            \textit{Answer:}
        \end{tcolorbox}
    \end{minipage}
    
\end{table*}

\begin{table*}[t]
    \centering
    
    \caption{RAG Prompt template for multiple-choice questions.}
    \label{tab:prompt5}    
    \begin{minipage}{0.95\columnwidth}
        \centering
        \begin{tcolorbox}[title=Multi-choice RAG Prompt]
            Answer the question based on the given document.
            
            Only give me the answer and do not output any other words.

            The following are given documents.

            \{ reference \}

            \textit{Question: \{ Instruction \}}
            
            \textit{Choices: \{ Choices \}}

            Make prediction from the given choices.
            
            \textit{Answer:}
        \end{tcolorbox}
    \end{minipage}
    
\end{table*}

\begin{table*}[t]
    \centering
    
    \caption{Prompt template for numerical questions.}
    \label{tab:prompt6}    
    \begin{minipage}{0.95\columnwidth}
        \centering
        \begin{tcolorbox}[title=Numerical RAG Prompt]
            Answer the question based on the given document.

            \textbf{Conclude the answer by stating ``The answer is therefore [ANSWER]``}
            
            Only give me the answer and do not output any other words.

            The following are given documents.

            \{ reference \}

            \textit{Question: \{ Instruction \}}
            
            \textit{Answer:}
        \end{tcolorbox}
    \end{minipage}
    
\end{table*}

\begin{table*}[t]
    \centering
    
    \caption{Prompt template for generation questions.}
    \label{tab:prompt7}    
    \begin{minipage}{0.95\columnwidth}
        \centering
        \begin{tcolorbox}[title=Generation RAG Prompt]
            Answer the question based on the given document.

            Your answer should be surrounded by [ANSWER] and [/ANSWER]. When generating a molecule, please generate a valid SMILES string.

            The following are given documents.

            \{ reference \}
            
            Only give me the answer and do not output any other words.

            \textit{Question: \{ Instruction \}}
            
            \textit{Answer:}
        \end{tcolorbox}
    \end{minipage}
    
\end{table*}

\end{document}